\pdfoutput=1

\documentclass[11pt]{article}

\usepackage[]{ACL2023}
\usepackage{xcolor}
\definecolor{darkgreen}{rgb}{0.0, 0.5, 0.0}
\definecolor{lightblue}{RGB}{173,216,230}
\definecolor{lightred}{RGB}{255,182,193}
\definecolor{lightgreen}{RGB}{173,255,47}
\definecolor{lightyellow}{RGB}{255,255,204}
\definecolor{violet}{RGB}{90, 19, 242}

\usepackage[most]{tcolorbox}
\usepackage{times}
\usepackage{latexsym}
\usepackage{graphicx}
\usepackage{subcaption}
\usepackage{hyperref}
\usepackage{wrapfig}
\usepackage{url}
\usepackage{graphicx}
\usepackage{multirow}
\usepackage{hyperref}
\usepackage{booktabs}
\usepackage{longtable}
\usepackage{longtable}
\usepackage{tabularx}
\usepackage{booktabs}
\usepackage{siunitx}
\usepackage{hyperref}
\usepackage{enumitem}
\usepackage{amsmath}
\usepackage{xcolor}
\usepackage{tcolorbox}
\usepackage{soul}
\usepackage{makecell}
\usepackage{multirow}
\usepackage{booktabs}
\usepackage{graphicx}
\usepackage{amsmath}
\usepackage{amssymb}
\usepackage{placeins}
\usepackage[titletoc]{appendix}
\usepackage{xltabular}
\usepackage{amsmath}
\usepackage{makecell}
\usepackage{colortbl}

\usepackage[T1]{fontenc}

\usepackage[utf8]{inputenc}

\usepackage{microtype}

\usepackage{inconsolata}

\tcbset{
    mybox/.style={
        colback=gray!10,    
        colframe=gray!50,   
        fonttitle=\bfseries, 
        boxrule=0.8mm,      
        title=#1            
    }
}

\title{Towards Context-Robust LLMs: \\A Gated Representation Fine-tuning Approach   }

\author{Shenglai Zeng$^{1}$, Pengfei He$^1$, Kai Guo$^1$, Tianqi Zheng$^{2}$ , \textbf{Hanqing Lu$^{2}$, Yue Xing$^1$, Hui Liu$^1$} \\ 
$^1$Michigan State University  \quad $^2$ Amazon.com   
  \\
\{zengshe1,xingyue1, liuhui7\}@msu.edu, \\
\{tqzheng, luhanqin\}@amazon.com
}

\begin{document}
\maketitle
\newtheorem{definition}{Definition}

\begin{abstract}
\label{abstract}
Large Language Models (LLMs) enhanced with external contexts, such as through retrieval-augmented generation (RAG), often face challenges in handling imperfect evidence. They tend to over-rely on external knowledge, making them vulnerable to misleading and unhelpful contexts. To address this, we propose the concept of context-robust LLMs, which can effectively balance internal knowledge with external context, similar to human cognitive processes. Specifically, context-robust LLMs should rely on external context only when lacking internal knowledge, identify contradictions between internal and external knowledge, and disregard unhelpful contexts. To achieve this goal, we introduce Grft, a lightweight and plug-and-play gated representation fine-tuning approach. Grft consists of two key components: a gating mechanism to detect and filter problematic inputs, and low-rank representation adapters to adjust hidden representations. By training a lightweight intervention function with only 0.0004\% of model size on fewer than 200 examples, Grft can effectively adapts LLMs towards context-robust behaviors.
\end{abstract}




\section{Introduction}
\label{Intro}


Providing large language models(LLMs) with external related contexts can improve their factual accuracy and reliability~\cite{gao2023retrieval, lewis2020retrieval,fan2024survey}. Techniques such as retrieval-augmented generation(RAG)~\cite{lewis2020retrieval} has gained widespread application including healthcare~\cite{amugongo2024retrieval}, legal services~\cite{wiratunga2024cbr}, and financial analysis~\cite{setty2024improving}.

However, this approach faces significant challenges when dealing with imperfect evidence. LLMs tend to over-rely on external knowledge, making them vulnerable to misleading or inaccurate information in the retrieved context. \citet{zou2024poisonedrag,dengpandora} demonstrated that contextual misinformation can mislead LLMs even when they possess the correct knowledge internally.  Besides, \citet{yoranmaking,fang-etal-2024-enhancing} demonstrate that a substantial portion of retrieved contexts, while semantically related to the query, can be unhelpful or irrelevant for answering the question, leading to degraded LLM performance.

\begin{figure}[t]
    \centering
    \includegraphics[width=1\linewidth]{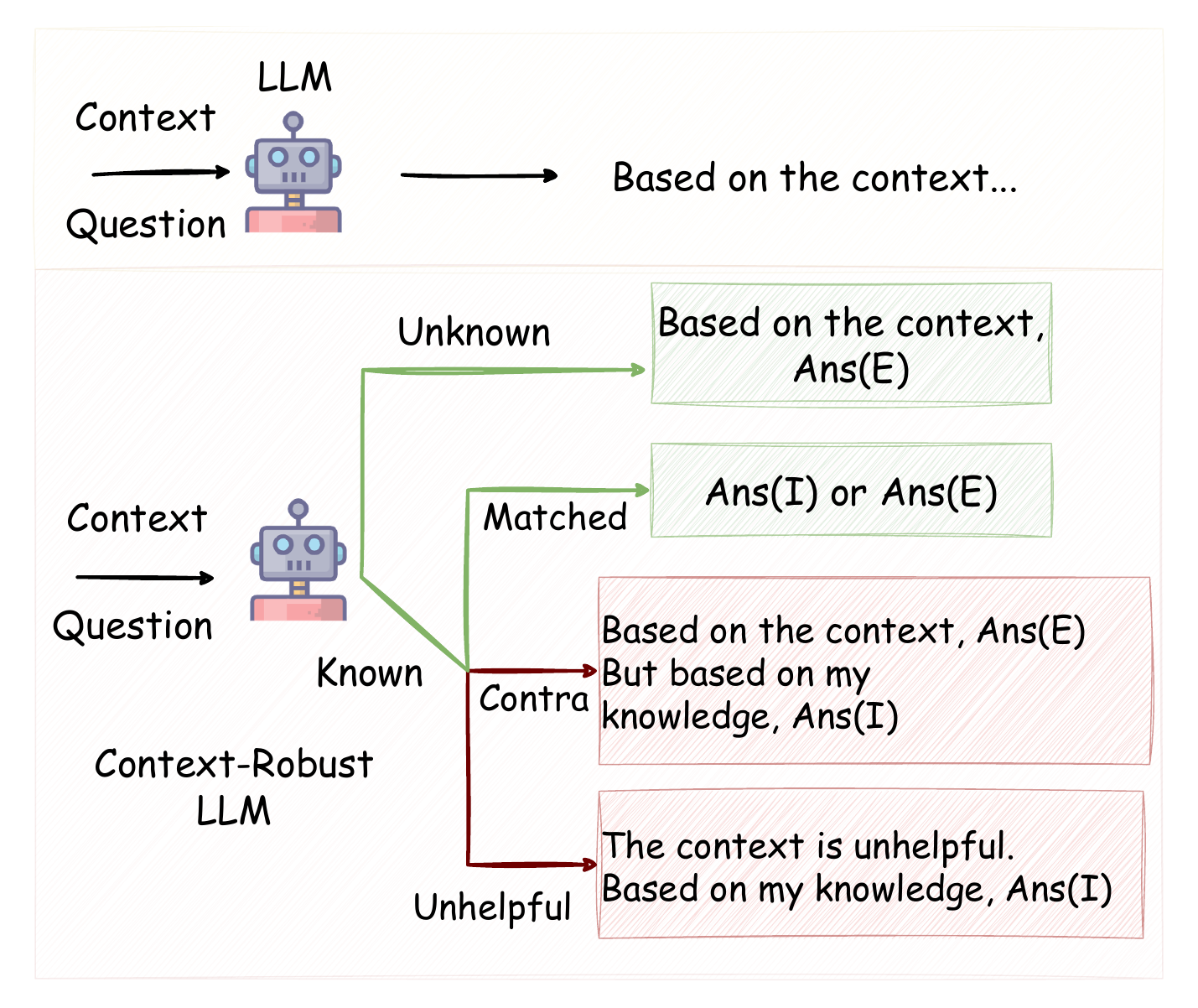}
    \caption{ Comparison between current LLMs and our developed context-robust LLMs in this work. Ans(I) refers to responses based on internal knowledge, while Ans(E) refers to responses based on external context. Current LLMs primarily rely on external sources for responses, whereas our context-robust LLM carefully balances contextual information with its internal knowledge to provide more reliable responses.}
    \label{fig:intro}
     \vspace{-0.2in}
\end{figure}

Compared to LLMs, humans demonstrate greater robustness when processing external information by carefully weighing it against their internal knowledge to reach reasoned conclusions \cite{hollister2017contextual}. This capability, often referred to as \textbf{contextual reasoning} or \textbf{knowledge integration}, is critical for ensuring reliable and accurate responses in real-world applications. For example, when presented with text claiming "Paris was the capital of France until 2020 when it moved to Marseille," people can reason "Based on my knowledge, Paris is France's capital, though this source claims it's Marseille." Similarly, when given irrelevant context about French cuisine while answering a capital city question, people naturally ignore the context and rely on their internal knowledge.  To match this capability, a \textbf{context-robust LLM} is desired to have similar cognitive processes as shown in Fig \ref{fig:intro}.

As discussed in previous works \cite{wang2024astute,yoranmaking}, LLMs can benefit from external contexts when they lack the knowledge to answer a question. This suggests an expected strategy for LLMs to utilize external contexts: \textbf{LLMs should rely on external context only when lacking internal knowledge} (the ``Unknown'' case in Fig \ref{fig:intro}). When internal knowledge exists, they can carefully balance external and internal knowledge to provide more objective and thoughtful answers\cite{chen2024benchmarking}. For example, as shown in Fig \ref{fig:intro}, when encountering knowledge that matches with its internal beliefs, a context-robust LLM will use both sources to formulate its response. While faced with contradictory evidence, it is to identify the contradiction and present both answers\footnote{Users can choose to use Ans(I) or Ans(E) based on their needs. For instance, for knowledge updating, they might choose Ans(E), while for addressing misinformation, they might prefer Ans(I).}. Similarly, if unhelpful contexts are presented, a context-robust LLM is expected to identify and ignore such contexts, relying solely on its internal knowledge to generate an answer\cite{yoranmaking}. However, current LLMs do not naturally exhibit these behaviors. Therefore, in this work, we aim to adapt LLMs' behaviors following the process in Fig~\ref{fig:intro} when handling retrieved-context.

Adapting LLMs to exhibit these behaviors presents tremendous challenges. Previous studies, such as \cite{zeng2024towards}, have shown that training-free methods—such as adding system prompts, in-context learning, or using Chain-of-Thought (CoT) prompting to guide models in balancing internal and external contexts—are far from reliable. Another approach is to fine-tune the LLM to teach it the desired behavior. However, direct end-to-end training of model parameters typically requires extensive training time and large datasets, which is not always feasible in real world, and limited examples may be insufficient to teach the LLM towards context-robust behaviors as we show later in Table \ref{tab:query_results}. 

In contrast, we aim to develop both data-efficient and training-efficient methods that intrinsically adapt LLMs to context-robust behaviors. \citet{zeng2024towards} show that LLMs' representations exhibit intrinsic distinct patterns when processing contradictory, matched, helpful, or unhelpful inputs. Furthermore, \citet{zou2023representation,wuandarora2024reft} demonstrate that intervening in LLMs' hidden layer representations can effectively modify their behavior patterns and improve task performance using only a few samples. These findings motivate us to leverage  representation adaptation to achieve the desired behaviors. It is particularly suitable to achieve our goal because it intrinsically relates to how the model processes external contexts\cite{zeng2024towards} (e.g., contradictory, unhelpful, or matched), enabling precise control over behavior. Additionally, it provides efficient, lightweight adaptation with minimal computational overhead and data requirements\cite{wu2024reft}.

In our work, we propose a gated representation fine-tuning pipeline Grft to enhance LLMs' robustness to retrieved contexts. Grft introduces a \textbf{lightweight, plug-and-play} intervention for LLMs' hidden layer representations. It consists of two components: (1) a gate mechanism that detects "abnormal" inputs and determines if intervention is needed, and (2) low-rank representation adapters that adjust hidden representations within a linear subspace. Using fewer than 200 training questions, we end-to-end train Grft (0.0004\% of model parameters). Experimental results demonstrate that Grft effectively improves LLM performance with misleading and unhelpful contexts while maintaining performance on helpful contexts.

\section{Related Work}
\label{Intro}

{\bf Robustness issues of RAG.} Although integrating external contexts is a common practice to enhance the quality and relevance of Large Language Models (LLMs), recent studies have revealed significant drawbacks associated with this approach. Research by \citet{ren2023investigating,wang2023resolving,ni-etal-2024-llms,liu2024ra,wang2023self,asaiself} highlights that current LLMs struggle to accurately assess the relevance of questions and determine whether retrieval is necessary. Additionally, studies such as \cite{zeng2024good, zou2024poisonedrag, dengpandora, xieadaptive} indicate that LLMs tend to over-rely on external contexts, even when these contexts conflict with their internal knowledge. Furthermore, the inclusion of unhelpful or irrelevant contexts can significantly degrade LLM performance, as noted by \cite{yoranmaking,fang-etal-2024-enhancing,chen2024benchmarking,sawarkar2024blended,wang2024astute,zeng2024towards,liu2024ra,zhao2024retrieval}. 


{\bf Representation Engineering and fine-tuning on LLMs.}
 Recently, a line of research indicates that LLMs' hidden representations contains rich information and can be utilized to efficiently modify model behaviors. For example, \cite{zou2023representation,zeng2024towards,lin2024towards,zheng2024prompt} demonstrates that the representations of large language models (LLMs) exhibit distinct patterns when processing contrasting concepts such as honesty and dishonesty, harmful and harmless, helpful and helpfulness.
 Besides, \citet{zou2023representation} also show the principal directions in low-dimensional spaces derived from these representations can be leveraged to control the behavior of LLMs.  Recently, \citet{wu2024reft} proposed ReFT, a technique to train interventions on LLM representations, enabling efficient control of model behavior for downstream tasks with minimal parameters.

\section{Method}
\begin{figure}[t]
    \centering
    \includegraphics[width=1\linewidth]{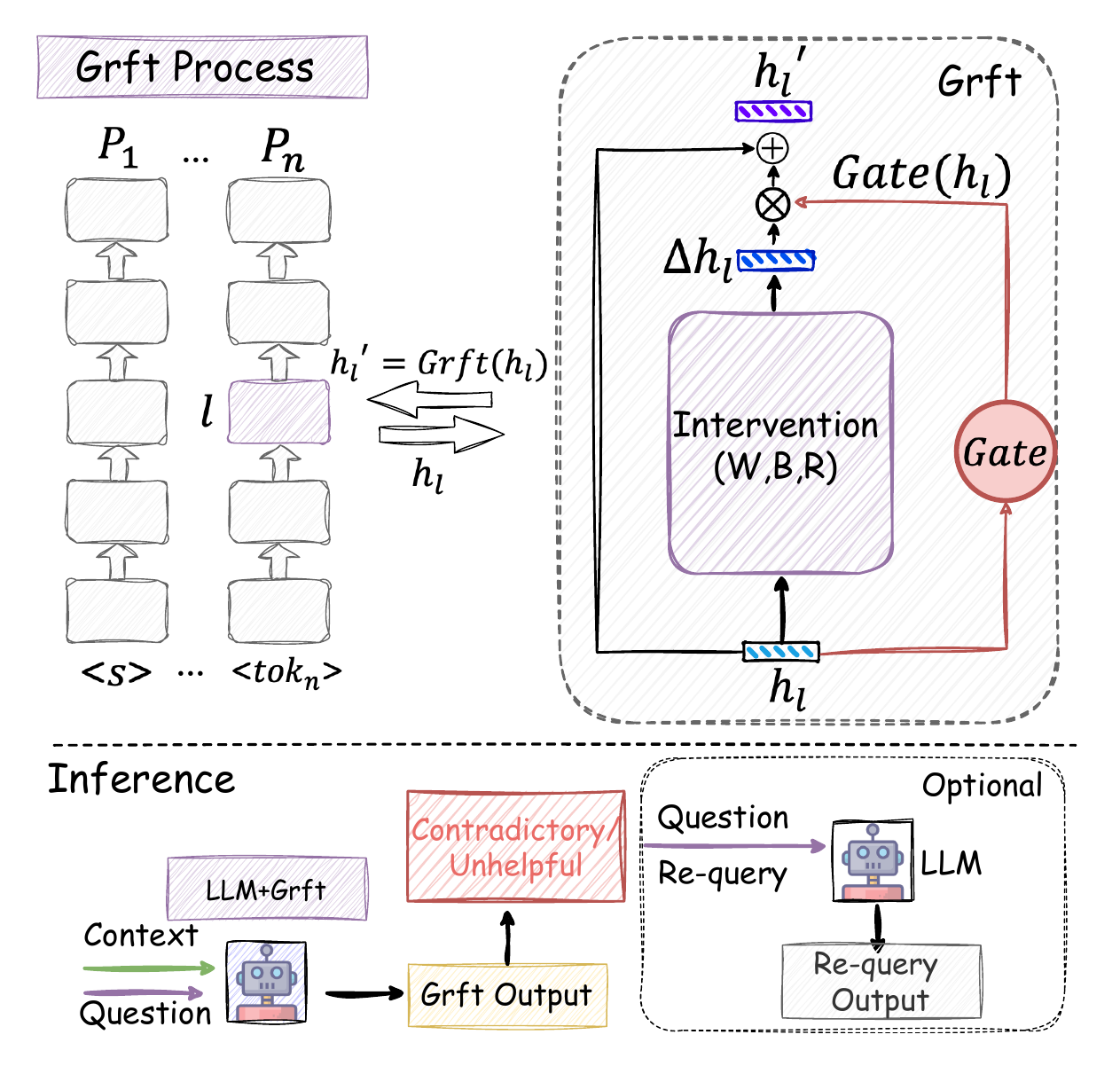}
    
    \caption{An overview of Grft}
    \label{fig:grft}
     \vspace{-0.2in}
\end{figure}
\label{Sec: Representation}

In this section, we present our Grft design. Section \ref{sec:ps} introduces the Problem Setting, followed by an overview of Grft in Section \ref{sec:Overview}. We then detail Grft's two main components: the gate function (Section \ref{Sec:gate}) and intervention (Section \ref{Sec: Inter}). Finally, we describe parameter updating (Section \ref{Sec: para_up}) and the inference (Section \ref{Sec:Inference}).
\subsection{Problem Setting}
\label{sec:ps}

Our goal is to obtain an \textbf{context-robust LLM} which can generate \textbf{context-robust responses} as illustrated in Fig \ref{fig:intro}. In particular, we expect the context-robust LLM satisfies the following:

\begin{itemize}[noitemsep,topsep=0pt] 
    \item[(a)] Rely on external information when it lacks internal knowledge.  
    \item[(b)] Use either internal or external knowledge when they match. 
    \item[(c)] Identify and resolve contradictions by providing both answers when external context conflicts with internal knowledge.  
    \item[(d)] Ignore unhelpful contexts and rely solely on internal knowledge when necessary.  
\end{itemize}

To achieve this, our method utilizes training data \( s_i = \{x_i, y_i, z_i\} \), where each sample consists of:  
\begin{itemize}[noitemsep,topsep=0pt] 
    \item An input \( x_i = \{c_i \parallel q_i\} \), where \( c_i \) is the given context (contradictory, unhelpful, matched, or helpful) and \( q_i \) is the question (known or unknown to the LLM);  
    \item A desired output \( y_i \) that follows the logic of a context-robust LLM, as described above;  
    \item A gate label \( z_i \) indicating whether intervention is necessary. For cases (a) and (b), \( z_i = 0 \) since the original LLM already exhibits the desired behavior. For cases (c) and (d), \( z_i = 1 \) as the LLM does not naturally produce the required behavior.  
\end{itemize}








To achieve the above goal, we propose and train an intervention function, \textbf{Grft($\cdot$)}, on the internal representations of LLMs, so that the LLM can better distinguish matched/contradictory/unhelpful/helpful contexts as in Fig.~\ref{fig:intro}. In the following, we introduce the details of the proposed Grft. 

\subsection{An Overview of Grft}
\label{sec:Overview}


As shown in Fig.~\ref{fig:grft}, in the Grft \textbf{training stage}, we aim to train a Grft intervention function that is composed of 2 components: a \textbf{gate} function that evaluates the input $h_l$ and decides whether the input context needs representation intervention, and an \textbf{intervention} component that projects and intervenes LLMs' representation in a low rank space.  The overall intervention function can be expressed as follows:
\begin{equation}
    \text{Grft}(\mathbf{h}_l) = \mathbf{h}_l + \text{Gate}(\mathbf{h}_l) \cdot \text{Intervention}(\mathbf{h}_l)
    \label{eq:grft}
\end{equation}
In the following subsections, we will provide a detailed explanation of each component including $\text{Gate}(\mathbf{h}_l)$ and $\text{Intervention}(\mathbf{h}_l)$.


While in the \textbf{inference stage}, our method introduces two approaches for LLM+Grft: Grft directly generates robust outputs using Grft interventions, while Grft-requery enhances reliability by re-querying the LLM when outputs indicate contradictions or unhelpful contexts.


\subsection{Gate Function} 
\label{Sec:gate}
The \textbf{Gate function} is designed to evaluate whether the context is abnormal and potentially requires intervention. It takes the hidden representation \( \mathbf{h}_l \)  of the LLM as input and outputs a scalar value ranging from 0 to 1, which controls the degree of intervention. In the context of \textbf{contextual question answering}, as illustrated in Fig.~\ref{fig:intro}, we consider inputs to be ``normal'' when the LLM itself lacks knowledge about the input query and the provided context matchs with the LLM's internal knowledge. In such cases, the LLM's inherent behavior is sufficient to produce the correct answer, and we expect the Gate function to output a low value.  Conversely, if the LLM encounters a question for which it possesses knowledge but the external context is either contradictory or unhelpful, an intervention is necessary to ensure that the LLM exhibits \textbf{context-robust} behavior. In such scenarios, we expect the Gate function to produce a high value.  In our study, we primarily utilize the \textbf{sigmoid function} as the Gate function due to its ability to smoothly map inputs to the desired range of 0 to 1. The Gate Function is formally defined as:
\begin{equation}
    \text{Gate}(\mathbf{h}_l) = \sigma(\mathbf{W}_g \mathbf{h}_l + \mathbf{b}_g)
    \label{eq:gate}
\end{equation}
where \( \mathbf{h}_l \in \mathbb{R}^{d} \) is the input hidden representation with dimensionality \( d \), \( \mathbf{W}_g \in \mathbb{R}^{1 \times d} \) is a learnable weight matrix, \( \mathbf{b}_g \in \mathbb{R} \) is a learnable bias term, and \( \sigma(\cdot) \) is the Sigmoid activation function.


The output \text{Gate}$(\mathbf{h}_l)$ serves as a gating signal: when it is close to 1, a strong intervention is required, while a value close to 0 indicates that little to no intervention is needed, allowing the model to rely primarily on its intrinsic behavior. This mechanism enables an adaptive balance between the model's original output and external interventions, ensuring robust performance across diverse contexts.


\subsection{Intervention} 
\label{Sec: Inter}
The intervention component aims to learn an intervention on LLMs' representations in low dimension space towards a more reliable answer. 
\begin{equation}
    \text{Intervention}(\mathbf{h}_l) = \mathbf{R}^\top(\mathbf{W}\mathbf{h}_l + \mathbf{b} - \mathbf{R}\mathbf{h}_l)
    \label{eq:delta}
\end{equation}
where $\mathbf{W}$ and $\mathbf{R}$ are low-rank matrixs and $\mathbf{b}$ is a bias vector that matches the dimensionality of $\mathbf{W}\mathbf{h}_l$ . The above term is also used in original ReFT~\cite{wu2024reft} methods. It
implements a low-rank fine-tuning mechanism through dimensional projection.  The linear transformation $\mathbf{W}\mathbf{h}_l + \mathbf{b}$ maps the representation to a new space, while $\mathbf{Rh}_l$ performs a low-rank projection. Their difference is then projected back through $\mathbf{R}^\top$ before being multiplied by the gate value and added to the input. The key difference between \textbf{Grft} and \textbf{ReFT} lies in the introduction of the \textbf{Gate} ($\mathbf{h_l}$), which regulates the extent of intervention.

\subsection{Parameter Updatating}
\label{Sec: para_up}

\paragraph{Loss Function.} The final loss function consists of two components:  a standard supervised fine-tuning loss and a gate supervision loss:
\begin{equation}
    \mathcal{L}_{\text{total}} = \mathcal{L}_{\text{FT}}(\hat{y}_i,y_i ) + \mathcal{L}_{\text{gate}}(\text{Gate}(\mathbf{h}_l^i),z_i)
    \label{eq:loss}
\end{equation}
where $\mathcal{L}_{\text{FT}}$ is the standard cross-entropy loss between model outputs $\hat{y}_i$ and ground-truth $y_i$, and $\mathcal{L}_{\text{gate}}$ is computed using binary cross-entropy loss to supervise the gate values:
\begin{equation}
   \begin{aligned}
   \mathcal{L}_{\text{gate}} =  & z_i \log(\text{Gate}(\mathbf{h}_l^i)) \\ 
   & + (1-z_i) \log(1-\text{Gate}(\mathbf{h}_l^i)) 
   \end{aligned}
   \label{eq:gate_loss}
\end{equation}
where $z_i$ represents the binary label indicating whether intervention is needed for the $i$-th sample, and $B$ is the number of samples in the batch.  

\paragraph{Training.} The learnable parameters in Grft include the gating mechanism parameters ($\mathbf{W_g}, \mathbf{b_g}$) and the intervention process parameters ($\mathbf{W}, \mathbf{b}, \mathbf{R}$). During training, we freeze the base model parameters and only update these learnable parameters through backpropagation. This approach introduces minimal computational overhead since these parameters constitute only a tiny fraction of the full model's parameter count.



\subsection{Grft Inference and Requery.}
\label{Sec:Inference}
As shown in Figure \ref{fig:grft}, we design two strategies to utilize the LLM with Grft interventions (LLM+Grft). The first strategy, denoted as \textbf{Grft}, directly prompts LLM+Grft with questions and contexts to generate more robust outputs. The second strategy, \textbf{Grft-requery}, focuses on reliable knowledge recall: when the Grft output contains indicators such as "CONTRADICTORY" or "UNHELPFUL," we re-query the original LLM and replace the internal answer \textbf{{ans(I)}} in the template "Based on what I know, \textbf{{ans(I)}}" with the \textbf{LLM's  answer}. This re-querying process is optional, as it requires querying the model twice. In  Section \ref{sec:experiment}, we compare this approach with other methods that also involve multiple queries during inference. 

\begin{figure*}[!t]
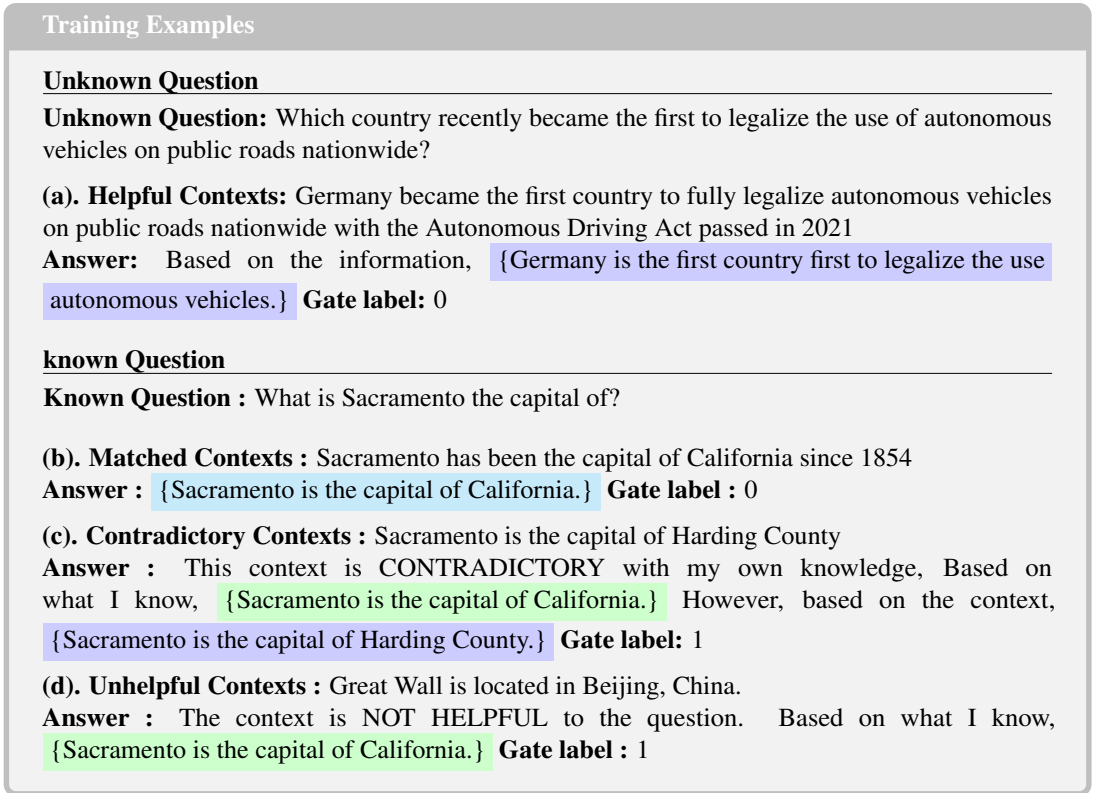

\vspace{-10 pt}
\centering
\resizebox{0.9\textwidth}{!}{
\begin{tcolorbox}[mybox={Training Examples}]
\textbf{Unknown Question}
\hrule
\medskip
\textbf{Unknown Question:} {Which country recently became the first to legalize the use of autonomous vehicles on public roads nationwide?}

\medskip

\textbf{(a). Helpful Contexts:} {Germany became the first country to fully legalize autonomous vehicles on public roads nationwide with the Autonomous Driving Act passed in 2021}

\textbf{Answer:} {Based on the information, \colorbox{blue!20}{\{Germany is the first country first to legalize the use}  \colorbox{blue!20} {autonomous vehicles.\}}}
\textbf{Gate label:} {0}

\bigskip
\textbf{known Question}
\hrule

\medskip

\textbf{Known Question :} {What is Sacramento the capital of?}

\medskip

\medskip

\textbf{(b). Matched Contexts :} {Sacramento has been the capital of California since 1854}

\textbf{Answer :}{ \colorbox{cyan!20}{\{Sacramento is the capital of California.\}}}
\textbf{Gate label :} {0}

\medskip

\textbf{(c). Contradictory Contexts :} {Sacramento is the capital of Harding County}

\textbf{Answer :} {This context is CONTRADICTORY with my own knowledge, Based on what I know, \colorbox{green!20}{\{Sacramento is the capital of California.\}} However, based on the context, \colorbox{blue!20}{\{Sacramento is the capital of Harding County.\}}}
\textbf{Gate label:} {1}

\medskip

\textbf{(d). Unhelpful Contexts :} {Great Wall is located in Beijing, China.}

\textbf{Answer :} {The context is NOT HELPFUL to the question. Based on what I know, \colorbox{green!20}{\{Sacramento is the capital of California.\}}}
\textbf{Gate label :} {1}

\end{tcolorbox}}
\vspace{-10 pt}
\caption{Training Examples.  }
\label{fig:training_samples}
\vspace{-10 pt}
\end{figure*}

\section{Experiment}\label{sec:experiment}

In this section, we conduct extensive experiments to validate the effectiveness of our methods. We present the results obtained from interventions performed on the 7th layer, as it demonstrates stable and satisfactory performance. Due to the space limitation, the ablation studies on different layer interventions and training sample requirements can be found in Appendix \ref{APP:ablation}.   

\subsection{Experimental Settings}
\label{sec:ex_setting}
\paragraph{Model and Baselines.}
In our experiments, we primarily employ the Llama-2-7B-Chat model as our main generation model, supplemented by results from Llama-3-8B-Instruct, which are detailed in Appendix \ref{app:llama3}. 

We conduct comprehensive comparisons of our methods against various training-based and prompting-based approaches. For training-based methods, we compare our approach with both full fine-tuning and LoRA fine-tuning, using the same training dataset. For prompting-based methods, we evaluate our methods against three commonly used strategies: (1) incorporating system instructions to explicitly guide the model in balancing external knowledge with its internal beliefs, (2) utilizing zero-shot Chain-of-Thought (CoT) prompting to encourage step-by-step reasoning for more thoughtful responses \cite{wei2022chain}, and (3) applying in-context learning by providing the LLM with examples that yield more reliable answers \cite{}. Additionally, we compare our methods with Astute RAG, a technique that involves multi-round prompting to help the LLM elicit and select between internal and external knowledge to answer questions \cite{wang2024astute}. Besides, we also report the performance of Grft without gate function(Grft-W/O Gate), and Grft with gate function but does not involve gate loss in training(Grft-W/O training). More details on these baseline methods are provided in Appendix \ref{App: baselines}.


\paragraph{Dataset.} We primarily utilize a subset of \href{https://github.com/OSU-NLP-Group/LLM-Knowledge-Conflict/tree/main}{ConflictQA} \cite{xieadaptive} as this benchmark provides both matched and contradictory evidence and answer to each question. Each sample consists of a \textit{PopQA} \cite{mallen-etal-2023-trust} question, the correct short and long evidence matched with the question, and ChatGPT-generated contradictory long and short evidence. To determine whether the LLM knows the question or not, i.e., Known/Unknown in Figure~\ref{fig:training_samples}, for each $q_i$, we prompt the LLM three times. If the LLM correctly answers the question in all three attempts, we classify the $q_i$ as a Known question. Conversely, if the LLM fails to provide the correct answer in all three attempts, we assume that the LLM lacks knowledge about the question and classify it as an unknown question. We obtain 5587 unknown questions and 1391 known questions for Llama-2-Chat-7B. Besides, for each known question, we randomly select one right matched context from other questions as the unhelpful random context and retrieve several pieces of evidence from the Wikipedia database, selecting the one with the highest retrieval score that does not contain the correct answer  as the unhelpful distracted context. We randomly selected 100 known questions and 100 unknown questions for training. 

\paragraph{Training Data Construction.} 
As illustrated in the training examples in Figure \ref{fig:training_samples}, there are four types of training samples. (a). \textbf{Unknown pairs:} If the LLM lacks knowledge  (Unknown Question in Figure \ref{fig:training_samples}), the gate label \( z_i \) is 0, and the output \( y_i \) corresponds to the correct answer. We randomly choose one of the short or long right contexts provided in the benchmark as unhelpful context $c_i$. (b).\textbf{Matched samples:}  When the context matches with its own knowledge(Known Question $\Rightarrow$ Matched Contexts in Figure \ref{fig:training_samples}). The gate label \( z_i \) is set to \( 0 \), and the ground truth output \( y_i \) corresponds to the correct answer to the question. We randomly choose one of the short and long matched contexts provided by \cite{xieadaptive} as matched context $c_i$. (c) \textbf{Contradictory samples:} if the context \( c_i \) contradicts the LLM's internal knowledge (Known Question $\Rightarrow$ {Contradictory} in Figure \ref{fig:training_samples}),  The gate label \( z_i \) is set to \( 1 \), $y_i$ follows the following template: identify the conflict  and provide an objective answer that combines  internal knowledge \colorbox{green!20}{\{ans(I)\}} and external knowledge \colorbox{blue!20}{\{ans(E)\}}. We utilize the ground truth output as {\{ans(I)\}} and the external evidence based answer provided in the  benchmark as {\{ans(E)\}}. We randomly choose one of the short and long contradictory contexts provided in the benchmark as contradictory context $c_i$. (d). \textbf{Unhelpful pairs: } if the LLM knows the answer  but the context  is unhelpful (Known Question $\Rightarrow$ {Unhelpful} in Figure \ref{fig:training_samples}), the gate label \( z_i \) is set to \( 1 \), and the output \( y_i \) should indicate the context's lack of usefulness and respond based on the model's own knowledge \colorbox{green!20}{\{ans(I)\}}.  We utilize the ground truth output as {\{ans(I)\}}. We randomly choose one of the random and distracted contexts provided in the benchmark as unhelpful context $c_i$. In the training process, we choose the rank $r=4$, batch size $B=5$ and optimize for 100 rounds.

\subsection{Main Results}
In this subsection, we evaluate the performance of Grft and baselines. For known queries, we test model accuracy under different conditions by providing contradictory (long and short), unhelpful (random and distracted), and aligned (long and short) contexts alongside the question. For unknown queries, we evaluate performance using helpful (long and short) contexts. We categorize contradictory and unhelpful contexts as ``noisy inputs'' since they can harm LLM performance, while aligned and helpful contexts are labeled as ``normal inputs'' as they do not degrade performance. 

\begin{table*}[htbp]
\centering
\caption{Results on Different Query Types (\%). Intervention is conducted on the 7-th layer.}
\label{tab:query_results}
\resizebox{0.8\textwidth}{!}{%
\begin{tabular}{l|ccccc|cc}
\hline
\multirow{3}{*}{Method} & \multicolumn{5}{c|}{Known queries} & \multicolumn{2}{c}{Unknown queries} \\
\cline{2-8}
 & \multicolumn{2}{c}{\textcolor{magenta}{Contradictory}} & \multicolumn{2}{c}{\textcolor{magenta}{Unhelpful}} & \multirow{2}{*}{\textcolor{olive}{Matched}} & \multicolumn{2}{c}{\textcolor{olive}{Helpful Context}} \\
\cline{2-5}\cline{7-8}
 & \makecell{\textcolor{magenta}{Short}} & \makecell{\textcolor{magenta}{Long}} & \makecell{\textcolor{magenta}{Random}} & \makecell{\textcolor{magenta}{Distracted}} & & \makecell{\textcolor{olive}{Short}} & \makecell{\textcolor{olive}{Long}} \\
\hline
LLM & 34.55 & 25.33 & 53.14 & 44.62 & 99.26 & 97.27 & \textbf{97.09} \\
ICL & 21.07 & 23.01 & 22.77 & 24.94 & 80.79 & 80.81 & 92.00 \\
CoT & 41.83 & 36.02 & 42.68 & 36.25 & 99.04 & \textbf{98.32} & 95.23 \\
System Prompt & 35.48 & 25.56 & 53.68 & 44.85 & 91.87 & 96.85 & 95.72 \\
FT-Llama-Lora & 31.68 & 27.42 & 52.21 & 47.25 & 97.68 & 95.52 & 89.79 \\
FT-Llama-Full & 32.68 & 26.94 & 54.08 & 47.28 & 95.29 & 96.29 & 93.08 \\
\rowcolor{gray!20} Astute-RAG & 59.60 & 46.70 & 72.56 & 69.80 & 93.60 & 72.88 & 86.73 \\
\hline
Grft-W/O Gate & 60.42 & 45.39 & 72.99 & 67.70 & 94.13 & 89.03 & 88.36 \\
Grft-W/O Loss & 41.05 & 36.48 & 72.30 & 68.09 & 98.14 & 92.36 & 90.17 \\
Grft & 60.88 & 61.19 & 73.22 & 68.86 & 99.07 & 98.23 & 97.03 \\
\rowcolor{gray!20} Gtft-requery & \textbf{82.49} & \textbf{88.15} & \textbf{97.68} & \textbf{98.46} & \textbf{99.38} & 97.30 & 96.99 \\
\hline
\end{tabular}%
}
\vspace{-0.1in}
\end{table*}
\subsubsection{Performance on abnormal inputs}

 
\paragraph{Contradictory contexts.} As shown in Table \ref{tab:query_results}~(column ``Contradictory"), we first observe that providing LLMs with contradictory evidence significantly harms performance. Even when the LLM itself has the correct answer, the accuracy drops to only 34.55\% (short) and 25.33\% (long). Additionally, while some training-free methods slightly improve performance (e.g., 41.83\% (short) and 36.02\% (long) with CoT), the results remain far from reliable. Moreover, training-based methods (FT-Full and FT-LoRA) do not show substantial improvements over the original LLM. This may be because directly updating the LLM's parameters with a limited number of samples is insufficient to teach the model the desired behavior. 

In contrast to the baseline methods, our methods show superior effectiveness, with Grft achieving the highest performance: 60.88\% (short contexts) and 61.19\% (long contexts), outperforming the original LLM by 26.33\% and 34.86\%, respectively. This highlights Grft's ability to teach the LLM more robust answering behavior, some cases are shown in Appendix \ref{App:grft_example}
 Re-querying further boosts performance to 82.49\% and 88.15\%, significantly exceeding Astute RAG, which also queries the model multiple times. These results suggest that Grft enables the LLM to effectively detect contradictions between internal and external answers.

\paragraph{Unhelpful contexts.} As in Table \ref{tab:query_results}~(column ``Unhelpful"), when LLMs are provided with random contexts (irrelevant to the query) or distracted contexts (superficially relevant but lacking the correct answer), we observe significant performance degradation. Even when the LLM itself possesses the correct knowledge, accuracy drops to 53.14\% (random) and 44.62\% (distracted). Training-free methods, such as ICL and CoT, prove ineffective, with system prompts yielding less than 1\% improvement. Similarly, training-based baselines show minimal gains (less than 3\%), highlighting the limitations of directly updating model parameters.

Compared to the other methods, our method demonstrate substantial improvements, with Grft achieving 73.22\% (+20.08\%) on random contexts and 68.86\% (+24.24\%) on distracted contexts, validating its effectiveness. After re-querying, performance rises to 97.98\% and 98.46\%, surpassing Astute RAG by 25.12\% and 29.66\%. These results indicate that Grft nearly identifies all unhelpful queries, though it may occasionally fail to recall its own answer without re-querying.

\subsubsection{Performance on normal inputs}

\paragraph{Matched contexts.} When providing LLMs with contexts that matched with their existing knowledge, both external and internal data can yield high accuracy. As shown in Table \ref{tab:query_results}~(column "Matched"), LLMs achieve near-perfect accuracy (99.26\%) when provided with matched contexts. Other baselines also demonstrate satisfactory performance, although we observe some degradation with methods such as system prompts, Chain-of-Thought (CoT), and Astute-RAG. This suggests that these methods may inadvertently mislead LLMs in such cases. Additionally, Grft-W/O gate exhibits a 5\% performance degradation, likely due to unnecessary interventions being applied to the representations. In contrast, our Grft and Grft-requery methods maintain high performance, achieving 99.07\% and 99.38\% accuracy, respectively. This underscores the effectiveness of our gate design in preserving performance while avoiding unnecessary interventions.

\paragraph{Unknown queries with helpful contexts.}When providing questions that the LLM cannot answer on its own, helpful contexts enable the LLM to leverage external information effectively. In such cases, we expect the LLM to achieve high accuracy. As shown in Table \ref{tab:query_results}(column "(column ``Contradictory")"), the LLM achieves excellent performance (97.27\% for short and 97.09\% for long contexts) when the correct context is provided. However, some baselines exhibit significant performance degradation. For instance, ICL achieves only 80\% accuracy on helpful short contexts, likely because the LLM is misled by in-context examples and fails to fully utilize external evidence. Similarly, FT-LoRA and FT-Full show reduced accuracy (89.79\%, -5.30\% and 93.08\%, -4.01\%, respectively) on helpful long contexts, indicating that direct fine-tuning can impair the LLM's ability to leverage helpful contexts. Astute RAG performs the worst, suffering performance losses of 24.39\% (short) and 10.36\% (long), as it first prompts the LLM without context, introducing incorrect answers that degrade performance when external evidence is later provided. 

Among representation-based methods, Grft-W/O Gate and Grft-W/O loss improve performance on ``noisy" inputs but degrade performance on helpful contexts, achieving only 89.03\% (short) and 88.36\% (long) for Grft-W/O Gate, and 92.36\% (short) and 90.17\% (long) for Grft-W/O loss. This is because minimal intervention is needed for helpful contexts, and unnecessary interventions in these methods lead to performance degradation. In contrast, our Grft and Grft-requery methods maintain performance comparable to the original LLM, validating the effectiveness of our gate design, which distinguishes between normal and noisy inputs and avoids excessive intervention in such cases.
\subsection{Gate Value analysis}
\begin{figure}[htpb]
\centering
    \includegraphics[width=0.7\columnwidth]{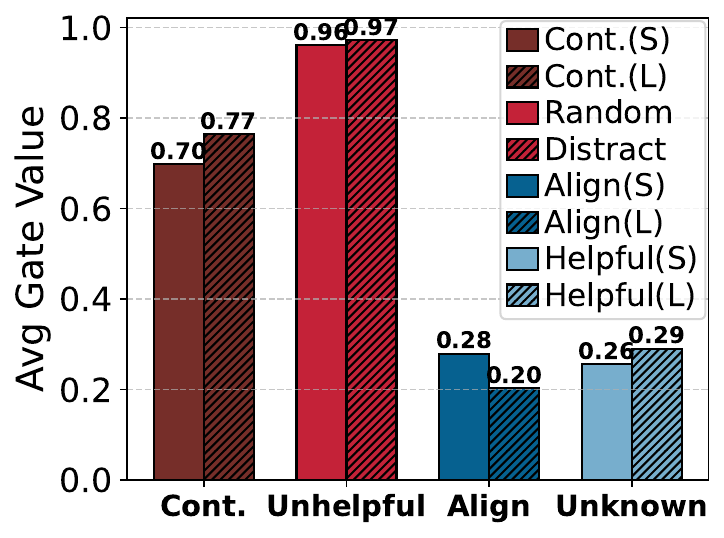}
\caption{Gate value on different contexts}
\vspace{-0.2in}
\label{Fig:gate_value}
\end{figure}

To validate the effectiveness of our gate design, we visualize the average gate values across all test queries for different contexts in Figure \ref{Fig:gate_value}. As shown, the gate value is significantly high for noisy inputs: it exceeds 0.7 for contradictory contexts and approaches 1 for unhelpful contexts with known queries. This demonstrates that the intervention mechanism is successfully activated for most noisy inputs. In contrast, for normal inputs—such as aligned contexts and unknown queries with helpful contexts—the average gate value remains below 0.3, indicating minimal intervention is applied to the representation, thereby preserving performance.

\begin{table}[htpb]
  \caption{Comparison of different fine-tuning methods.}
  \small
  \label{table:ft-comparison}
  \centering
  \renewcommand{\arraystretch}{1.2}
  \begin{tabular}{c|c|c} 
    \hline
    \textbf{Method} & \textbf{Parameters} & \textbf{Percentage} \\ 
    \hline\hline
    Full-FT & 6.74B & 100\% \\ 
    \hline
    LoRA (r=4) & 2.1M & 0.0311\% \\
    \hline
    ReFT(Grft-W/O Gate) (r=4) & 32.8K & 0.0005\% \\
    \hline
    GrFT (r=4) & 36.9K & 0.0005\% \\
    \hline
  \end{tabular}
  \vspace{-0.5cm}
\end{table}

\subsection{Parameters Comparison}
We compare the trainable parameters of our Grft method with baseline methods in Table \ref{table:ft-comparison}. Grft demonstrates high parameter efficiency, utilizing only 0.0005\% of the parameters required by Full-FT and 1.6\% of those used by LoRA-FT. Compared to ReFT, our method introduces a minimal 4.1K additional parameters (for the gate function) while achieving significantly more stable performance, as evidenced in Table \ref{tab:query_results}. This parameter efficiency allows the adaptation to be conducted in a resource-effective manner.


\subsection{Generazation performance}

To validate the generalization ability of our methods, we report the performance of Grft and Grft-requery—trained on ConflictQA—against other baselines on unseen datasets. Specifically, we evaluate the generalization ability on contradictory contexts using COUNTERFACT~\cite{meng2022locating}, a distinct dataset. We selected a subset of 1,000 samples from COUNTERFACT that Llama-7B-Chat can answer correctly. For each sample, we provided both contradictory and correct contexts alongside the question to the model and measured its performance.  For the generalization on unhelpful contexts, we utilized a 1200 subset of NQ (Natural Questions) that Llama-7B-Chat can answer. We paired these questions with random contexts from other NQ questions and distracted contexts constructed by \citet{Cuconasu_2024}. 

As shown in Table \ref{Tab:generazation}, we observe that Grft consistently enhances LLMs' performance when handling contradictory and unhelpful inputs. On COUNTERFACT, Grft achieves an accuracy of 62.52\% on contradictory inputs, which is 20.28\% higher than using the LLM directly. Furthermore, Grft-requery improves accuracy to 75.78\%, outperforming Astute-RAG by 24.35\%. On the NQ dataset (where Llama-7B-Chat knows the answers), Grft demonstrates performance improvements of 9.71\% (distracted contexts) and 24.11\% (random contexts) compared to the original LLM. Additionally, Grft-requery achieves high accuracies of 83.13\% (unhelpful contexts) and 82.20\% (random contexts).  These results indicate that Grft effectively captures the intrinsic patterns of noisy inputs and exhibits strong transferability.

\begin{table}[t]
\centering
\caption{Results on Knowns.QA and NQ (\%)}
\label{Tab:generazation}
\resizebox{\linewidth}{!}{%
\begin{tabular}{l|cc|cc}
\hline
\multirow{2}{*}{Method} & \multicolumn{2}{c|}{Knowns.QA} & \multicolumn{2}{c}{NQ} \\
\cline{2-5}
& \textcolor{magenta}{Misleading} & \textcolor{olive}{Right} & \textcolor{magenta}{Unhelpful} & Random \\
\hline
LLM & 42.24 & 93.56 & 52.47 & 38.84 \\
ICL & 14.32 & 77.21 & 21.89 & 11.50 \\
CoT & 47.61 & 94.57 & 40.20 & 34.07 \\
System Prompt & 45.82 & \textit{96.18} & 55.62 & 49.15 \\
Astute-RAG & 51.43 & 87.94 & 60.39 & 68.74 \\
FT-Llama-Lora & 37.47 & 90.45 & 36.29 & 45.32 \\
FT-Llama-Full & 43.15 & 91.23 & 42.35 & 47.98 \\
ReFT-Llama-Gate & \textit{62.52} & 95.11 & \textit{62.18} & \textit{62.95} \\
\rowcolor{gray!20} ReFT-Gate-re-query & \textbf{75.78} & \textbf{97.61} & \textbf{83.13} & \textbf{82.20} \\
\hline
\end{tabular}%
}
\end{table}

\vspace{-0.1in}
\section{Conclusion}
\vspace{-0.1in}
\label{Conclusion}
In this paper, we present Grft, a lightweight gated representation fine-tuning approach to enhance LLMs' contextual robustness. Through training a lightweight intervention function (parameters only accounting for 0.0004\% of model size)  on fewer than 200 samples, Grft effectively adapts LLMs to exhibit context-robust behaviors: relying on external context only when necessary, identifying contradictions and give integrated answers, and ignoring unhelpful contexts. Experimental results demonstrate that Grft significantly improves LLMs' robustness to imperfect contexts while maintaining their original capabilities, providing a practical solution for real-world applications where handling imperfect evidence is crucial.



\section{Limitations}
In this work, we primarily focus on enhancing LLMs' context-robust performance when processing single context-question pairs. Our current approach demonstrates effectiveness in improving contextual understanding, though there remain promising directions for future exploration. Specifically, we aim to investigate more sophisticated relationships, both internal and external, between different contexts to further enhance model performance. While our experimental evaluation centers on Llama-2-7B and Llama-3-8B-instruct, we plan to extend our analysis to a broader range of models to validate the generalizability of our approach and identify potential model-specific optimizations.
\bibliography{anthology}

\clearpage

\appendix
\onecolumn
\section{Appendix}
\href{}{}
\subsection{ablation studies}
\label{APP:ablation}
\begin{figure*}[htpb]

\centering
\resizebox{\textwidth}{!}{%
    \begin{minipage}{\textwidth}
        \subfloat[Contradictory]{\includegraphics[width=.25\textwidth]{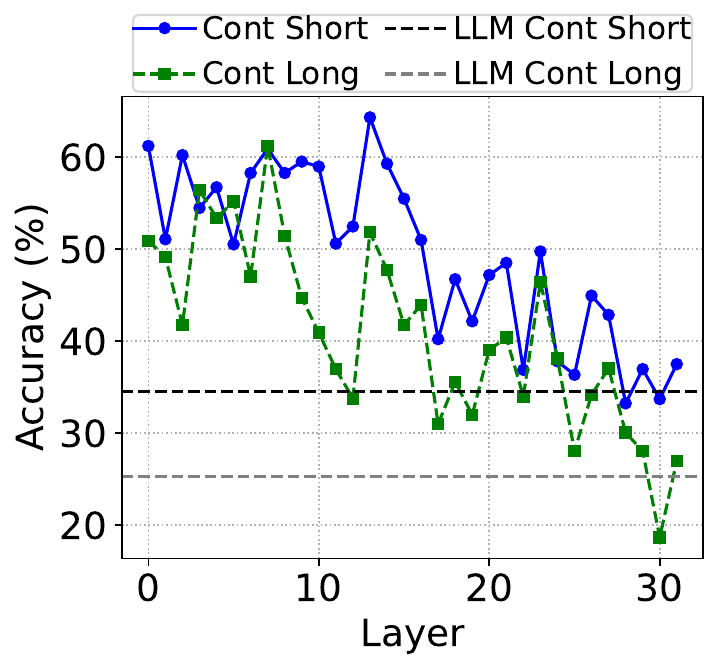}
        \label{fig:Layer Contra}}
        \subfloat[Unhelpful]{\includegraphics[width=.25\textwidth]{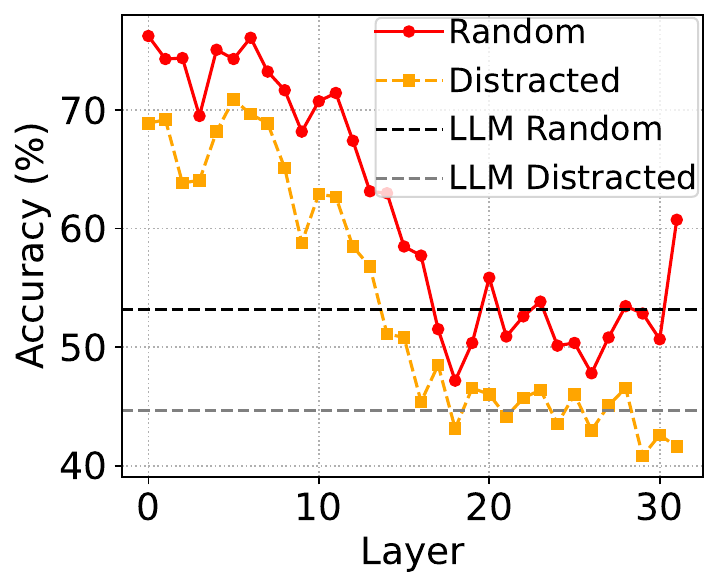}
        \label{fig:Layer unhelpful}}
        \subfloat[Aligned]{\includegraphics[width=.25\textwidth]{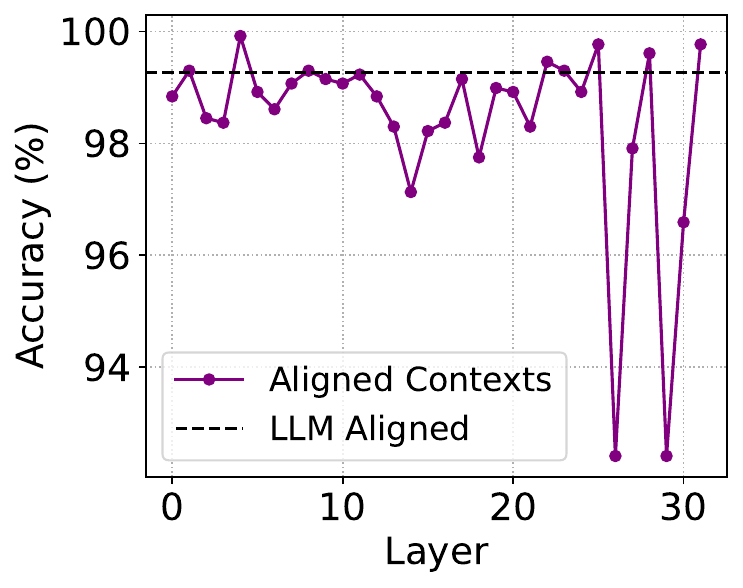}
        \label{fig:Layer Aligned}}
        \subfloat[Unknown]{\includegraphics[width=.25\textwidth]{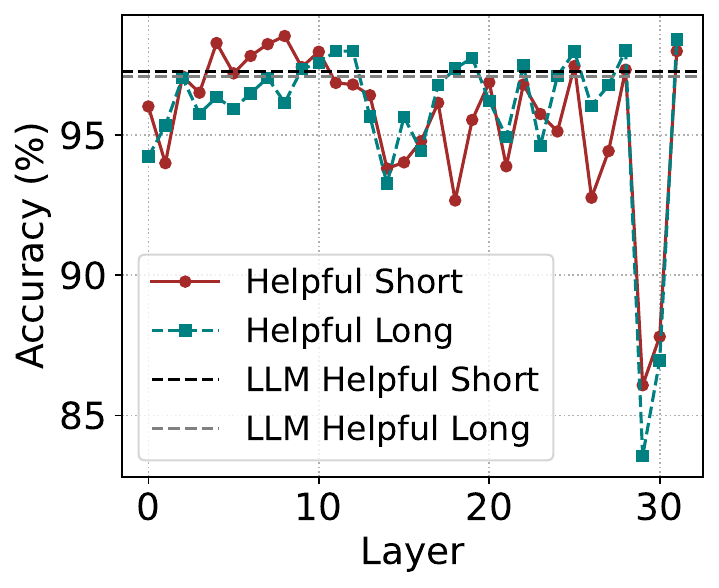}
        \label{fig:Layer unknown}}
    \end{minipage}
}
\caption{Ablation study on $i$-th layer intervention}
\label{Fig:ab_layer}
\end{figure*}

\begin{figure*}[htbp]

\centering
\resizebox{\textwidth}{!}{%
    \begin{minipage}{\textwidth}
        \subfloat[Contradictory]{\includegraphics[width=.25\textwidth]{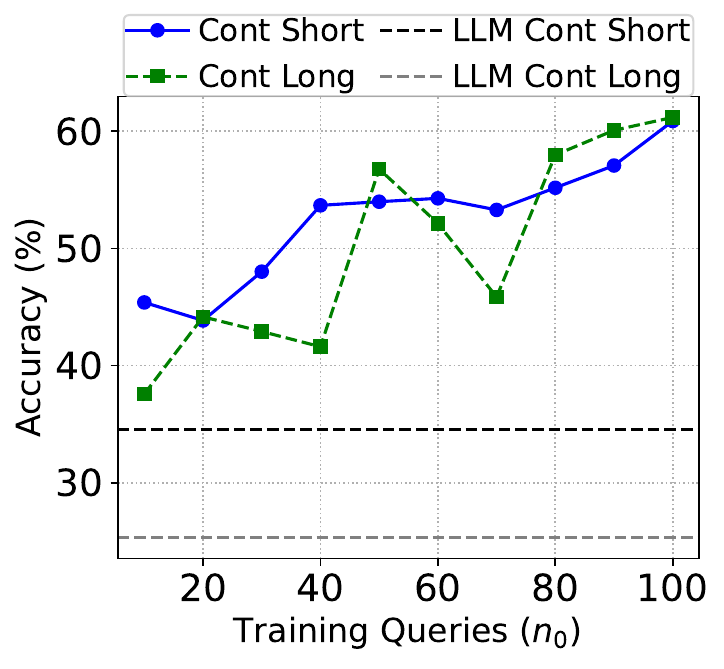}
        \label{fig:data_attributes_performance}}
        \subfloat[Unhelpful]{\includegraphics[width=.25\textwidth]{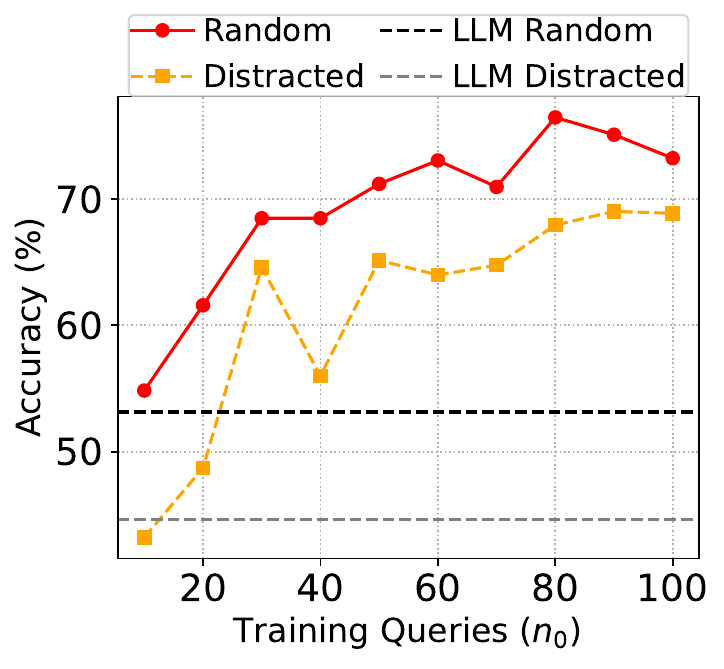}
        \label{fig:data_generator_performance}}
        \subfloat[Aligned]{\includegraphics[width=.25\textwidth]{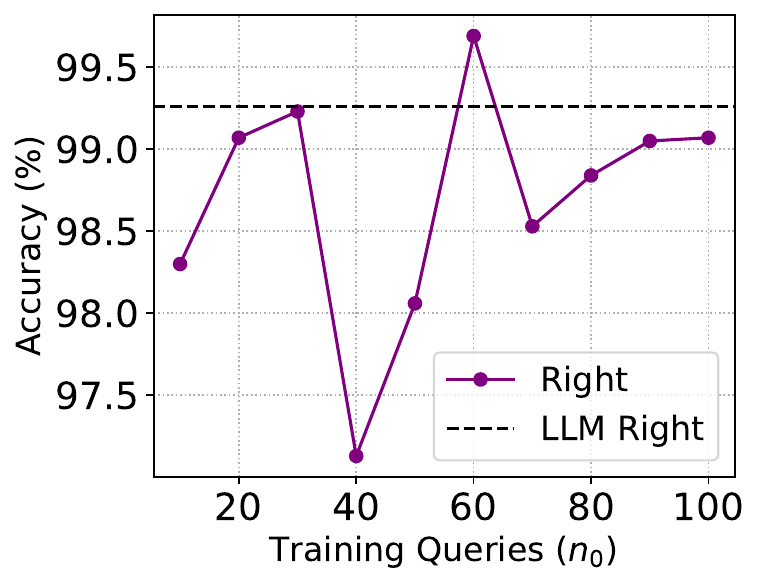}
        \label{fig:data_attributes_Attack}}
        \subfloat[Unknown]{\includegraphics[width=.25\textwidth]{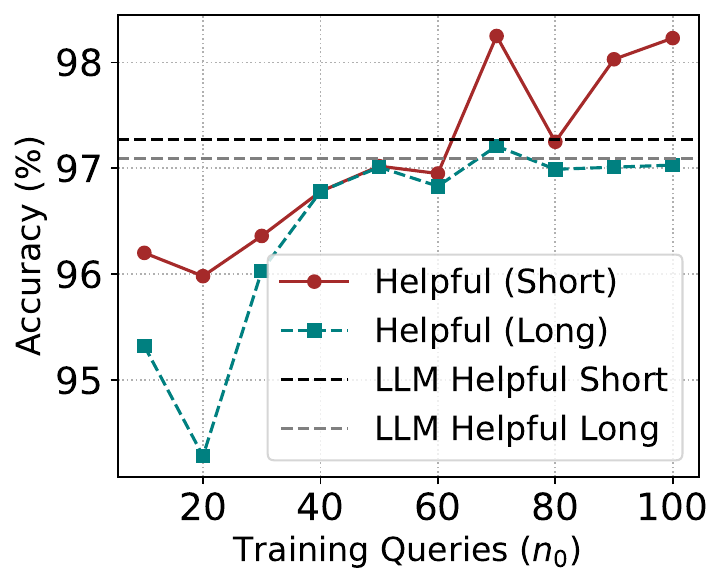}
        \label{fig:data_generator_Attack}}
    \end{minipage}
}
\caption{Ablation study on number of training queries}
\label{Fig:ab_sample}
\end{figure*}
In this section, we conduct ablation studies on the intervention effect on different layers and the minimal requirement of training samples.
\paragraph{Intervention on different layers.}
To explore which layers the intervention is effective to enhance the robustness performance, we plot the performance on test set under various situations(Contradictory, Unhelpful, aligned, unknown) with layers in Fig \ref{Fig:ab_layer}. As we can observe, doing intervention on the early layers is more effective (earlier than 15th layers). This may be because the internal knowledge is likely to be stored on middle layer MLPs\cite{meng2022locating}, and thus it's essential to change the representation in the early stage to help retrieve internal information. In contrast, if doing the intervention on later layers, the internal information is not likely to be retrieved and thus the performance on noisy query can not be effectively improved.
\paragraph{Training sample requirement.}
In this section, we investigate the minimal training data requirement to achieve reasonable performance. In our main experiments, we utilize \( N_1 = 100 \) known queries and \( N_2 = 100 \) unknown queries (totaling 200 queries and 400 samples) to train the intervention parameters. To explore the impact of reduced training data, we now vary the number of queries, using only \( N_1 = n_0 \) known queries and \( N_2 = n_0 \) unknown queries for training. We vary $n_0$ from 10 to 100.  The results are shown in Figure \ref{Fig:ab_sample}. 

We observe that even with fewer samples (e.g., \( n_0 = 60 \)), the model achieves stable and satisfactory performance. Furthermore, using as few as \( n_0 = 20 \) known and unknown training queries still improves performance compared to the original LLMs. These findings highlight the efficiency of our approach in leveraging limited training data to enhance model performance.

\subsection{Results on Llama-8B-Instruct}
\label{app:llama3}
In this section, we also present the results of Grft on the Llama-8B-Instruct model. We adhere to the experimental settings outlined in Section \ref{sec:ex_setting}. For Llama-8B-Instruct, we obtained 2,190 known and 4,429 unknown queries. We randomly sampled 100 known and 100 unknown queries to train the intervention functions on layer 0 (as it consistently delivers stable performance across all cases) and evaluated the performance on the remaining test set. As shown in Table \ref{tab:query_results_llama3}, both Grft and Grft-requery significantly improve the model's performance on noisy inputs while maintaining its effectiveness on matched contexts and unknown queries. This further validates the robustness and generalizability of Grft across different models.

\begin{table*}[htbp]
\centering
\caption{Llama-3 results.}
\label{tab:query_results_llama3}
\resizebox{0.8\textwidth}{!}{%
\begin{tabular}{l|ccccc|cc}
\hline
\multirow{3}{*}{Method} & \multicolumn{5}{c|}{Known queries} & \multicolumn{2}{c}{Unknown queries} \\
\cline{2-8}
& \multicolumn{2}{c}{\textcolor{magenta}{Contradictory}} & \multicolumn{2}{c}{\textcolor{magenta}{Unhelpful}} & \multirow{2}{*}{\textcolor{olive}{Matched}} & \multicolumn{2}{c}{\textcolor{olive}{Helpful Context}} \\
\cline{2-5}\cline{7-8}
& \makecell{\textcolor{magenta}{Short}} & \makecell{\textcolor{magenta}{Long}} & \makecell{\textcolor{magenta}{Random}} & \makecell{\textcolor{magenta}{Distracted}} & & \makecell{\textcolor{olive}{Short}} & \makecell{\textcolor{olive}{Long}} \\
\hline
LLM & 26.47 & 26.99 & 51.67 & 39.62 & 99.67 & 96.89 & 97.52 \\
ICL & 29.19 & 27.89 & 25.31 & 24.78 & 97.66 & 96.10 & 97.69 \\
CoT & 30.81 & 26.08 & 19.00 & 19.00 & 99.57 & 99.28 & 96.28 \\
System Prompt & 36.36 & 35.02 & 59.19 & 47.08 & 98.37 & 89.33 & 98.08 \\
\rowcolor{gray!20} Astute-RAG & 53.44 & 46.55 & 69.80 & 73.66 & 94.50 & 81.66 & 81.46 \\
FT-Llama-Lora & 32.08 & 30.18 & 26.09 & 25.89 & 93.28 & 94.32 & 96.03 \\
FT-Llama-Full & 31.98 & 29.02 & 27.19 & 24.96 & 95.03 & 95.26 & 94.39 \\
\hline
ReFT(Training) & 40.08 & 43.05 & 62.03 & 69.01 & 93.02 & 91.18 & 89.07 \\
ReFT-Gate-W/O loss & 39.33 & 44.07 & 63.06 & 68.85 & 95.90 & 92.09 & 91.05 \\
ReFT-Gate & 54.11 & 52.25 & 70.86 & 66.36 & 98.04 & 95.99 & 97.49 \\
\rowcolor{gray!20} ReFT-Gate-re-query & 69.47 & 78.18 & 82.02 & 96.03 & 99.71 & 95.38 & 97.51 \\
\hline
\end{tabular}%
}
\end{table*}

\begin{figure*}[!t]
\vspace{-10 pt}
\centering
\resizebox{0.9\textwidth}{!}{
\begin{tcolorbox}[mybox={Prompts}]

\textbf{System Prompt}
\hrule
\medskip
\textbf{System:} You are an AI assistant specialized in answering questions via a two-stage evaluation process. First check if you can answer the question with your knowledge. If uncertain, use the provided context. If you have relevant knowledge, evaluate the context: use both sources if aligned, explicitly state conflicting perspectives if they disagree, or ignore irrelevant context and answer from your knowledge alone.
\medskip

\textbf{User:} Context: \{context\}
Question: \{question\}

\bigskip
\textbf{In-Context Learning}
\hrule
\medskip
\textbf{System:} You are a helpful assistant.
\medskip

\textbf{User:} Here are some examples:

Case 1 - No Internal Knowledge:
Context: Eleanor Davis is a cartoonist who has published graphic novels like "The Hard Tomorrow".
Question: What is Eleanor Davis's occupation?
I don't have confident knowledge about Eleanor Davis, so I'll rely on the context: Eleanor Davis is a cartoonist who publishes graphic novels.

Case 2 - Contradicting Knowledge:
Context: Eleanor Davis works as a marketing manager for a cosmetic company in New York City.
Question: What is Eleanor Davis's occupation?
This context CONTRADICTS my knowledge. I know Eleanor Davis is a cartoonist and illustrator. However, the context claims she is a marketing manager.

Case 3 - Aligned Knowledge:
Context: Eleanor Davis is an American cartoonist and illustrator who creates comic works for both adolescent and adult audiences.
Question: What is Eleanor Davis's occupation?
The context ALIGNS with my knowledge - Eleanor Davis is a cartoonist and illustrator.

Case 4 - Irrelevant Context:
Context: Eleanor before her—Eleanor of Normandy, an aunt of William the Conqueror, lived a century earlier.
Question: What is Eleanor Davis's occupation?
The context is NOT HELPFUL. Based on my knowledge, Eleanor Davis is a cartoonist and illustrator.

Now please answer:
Context: \{context\}
Question: \{question\}

\bigskip
\textbf{Chain-of-Thought}
\hrule
\medskip
\textbf{System:} You are a helpful assistant.
\medskip

\textbf{User:} Context: \{context\}
Question: \{question\}

Think step by step:
1. Knowledge Check: Do I have reliable information about this topic in my internal knowledge? What specifically do I know?
2. Context Analysis: 
  - If I don't know: What information does the context provide to answer this question?
  - If I do know: Compare context with my knowledge for alignment or conflicts
3. Evaluation:
  - Does context match my knowledge? 
  - Does it contradict what I know?
  - Is it relevant to answering the question?
4. Response Strategy:
  - Unknown topic: Use context
  - Aligned knowledge: Use either source
  - Conflicting information: Present both perspectives
  - Irrelevant context: Use my knowledge only

\end{tcolorbox}}
\caption{Three prompting approaches with their respective system and user prompts.}
\label{fig:baseline_prompts}
\vspace{-10 pt}
\end{figure*}

\subsection{Baseline Details}
\label{App: baselines}
\paragraph{Prompts used for ICL, CoT and System prompts.} We detail our prompts for ICL, CoT and system prompts in Fig \ref{fig:baseline_prompts}.
\paragraph{Lora fine-tuning and Full finetuning} We fine-tuned Llama-2-7b-chat-hf using LoRA with configurations: rank=4, alpha=8, dropout=0.05, targeting q\_proj and v\_proj modules. The model was trained for 100 epochs with a batch size of 4, learning rate of 4e-4, and under bfloat16 precision. We fine-tuned Llama-2-7b-chat-hf using full-parameter tuning with learning rate of 1e-5, batch size of 1, and 100 epochs. Training optimizations include gradient accumulation steps of 8, gradient checkpointing, fused AdamW optimizer, and warmup ratio of 0.03. The model uses bfloat16 precision.

\paragraph{Astute-RAG.} The Astute-RAG approach posits that externally retrieved knowledge may contain irrelevant, misleading, or even malicious information, which could adversely affect the performance of LLMs. This method iteratively integrates internal and external knowledge, ultimately determining the final output of the LLMs based on the reliability of the information.

Specifically, this method contains three stages: generate initial context, consolidate knowledge, and generate final answer. The Astute-RAG approach initially extracts key internal information about the input question. The generated internal knowledge will be integrated with the retrieved external information, with all sources explicitly annotated. The initial context follows this structure:
"Own memory: \{internal knowledge\}$\setminus$n External Retrieval: \{retrieved knowledge\}". This initial context undergoes $t-1$ iterations in the consolidation stage. Each iteration generates a new context by leveraging the initial context and the last generated context. In the final stage, the generated contexts are used to produce the final answer with the highest credibility score. The prompt utilized is shown in Figure \ref{fig:astute_prompts}.

\begin{figure*}[!t]
\vspace{-10 pt}
\centering
\resizebox{0.9\textwidth}{!}{
\begin{tcolorbox}[mybox={Prompts}]

\textbf{Stage 1: Generate Internal Knowledge}
\hrule
\medskip
Generate a document that provides accurate and relevant information to answer the given question. If the information is unclear or uncertain, explicitly state ’I don’t know’ to avoid any hallucinations.\\ \\
Question: \{question\} Document \\

\bigskip
\textbf{Stage 2: Consolidate the Knowledge}
\hrule
\medskip
Task: Consolidate information from both your own memorized documents and externally retrieved documents in response to the given question. \\ \\
* For documents that provide consistent information, cluster them together and summarize the key details into a single, concise document.\\
* For documents with conflicting information, separate them into distinct documents, ensuring each captures the unique perspective or data.\\
* Exclude any information irrelevant to the query.\\
For each new document created, clearly indicate:\\
* Whether the source was from memory or an external retrieval.\\
* The original document numbers for transparency.\\ \\
Initial Context: \{initial context\}\\
Last Context: \{last context\} \\ 
Question: {question}\\
New Context:\\

\bigskip
\textbf{Stage 3: Generate Final Answer}
\hrule
\medskip
Task: Answer a given question using the consolidated information from both your own memorized documents and externally retrieved documents.\\ \\
Step 1: Consolidate information \\
* For documents that provide consistent information, cluster them together and summarize the key details into a single, concise document. \\
* For documents with conflicting information, separate them into distinct documents, ensuring each captures the unique perspective or data.\\
* Exclude any information irrelevant to the query \\
For each new document created, clearly indicate: \\
* Whether the source was from memory or an external retrieval. \\
* The original document numbers for transparency. \\ \\
Step 2: Propose Answers and Assign Confidence \\
For each group of documents, propose a possible answer and assign a confidence score based on the credibility and agreement of the information. \\ \\
Step 3: Select the Final Answer \\
After evaluating all groups, select the most accurate and well-supported answer. \\
Highlight your exact answer within <ANSWER> your answer </ANSWER>. \\ \\
Initial Context: \{initial\ context\} \\
Consolidated Context: \{last context\}\\
Question: \{question\} \\ 
Answer: \\

\end{tcolorbox}}
\caption{Prompts of Astute-RAG}
\label{fig:astute_prompts}
\vspace{-10 pt}
\end{figure*}

\subsection{Dataset description.}
In our experiments, we primarily utilize the ConflictQA dataset, with COUNTERFACT and NQ datasets for generalization studies. The ConflictQA dataset combines questions from \textit{PopQA} \cite{mallen-etal-2023-trust} with both aligned and contradictory evidence. Each sample contains a question paired with concise and detailed supporting evidence, as well as ChatGPT-generated contradictory evidence in both short and long forms.
The COUNTERFACT dataset \cite{meng2022locating} provides questions with matched and contradictory answers, making it suitable for evaluating model performance on contradictory contexts. For unhelpful context evaluation, we use a subset of the Natural Questions (NQ) dataset \cite{na} following \citet{cuconasu2024power}. For each NQ question, we select the highest-ranked distractor context and a randomly chosen golden passage from another question as unhelpful context.

\subsection{Grft Answer Examples}
\label{App:grft_example}
In Fig \ref{fig:grft_model_responses}, we show some output examples of Grft when encountering contradictory and unhelpful contexts, we can observe in these examples, that Grft successfully adapts LLM to our desired context-robust behaviors.

\begin{figure*}[!t]
\vspace{-10 pt}
\centering
\resizebox{0.9\textwidth}{!}{
\begin{tcolorbox}[mybox={Model Outputs}]
\textbf{Contradictory Context Responses}
\hrule
\medskip
\textbf{Question 1:} {What type of movie is Flightplan?}

\textbf{Response:} {This context CONTRADICTS my existing knowledge. Based on what I know, \colorbox{green!20}{Flightplan is a thriller film}. However, based on the context, \colorbox{blue!20}{Flightplan is movie action}.}

\textbf{Question 2:} {Where is Karahasanlı located?}

\textbf{Response:} {This context CONTRADICTS my existing knowledge. Based on what I know, \colorbox{green!20}{Karahasanlı is located in Turkey}. However, based on the context, \colorbox{blue!20}{Karahasanlı is in fr}.}

\textbf{Question 3:} {What is Old Persian?}

\textbf{Response:} {This context CONTRADICTS my existing knowledge. Based on what I know, \colorbox{green!20}{Old Persian is an ancient language that was spoken in ancient Persia, which is now modern-day Iran}. However, based on the context, \colorbox{blue!20}{Old Persian is in British North America}.}

\bigskip

\textbf{Unhelpful Context Responses}
\hrule
\medskip
\textbf{Question 4:} {What is Scooter Braun's occupation?}

\textbf{Response:} {The context is NOT HELPFUL to the question. Based on what I know, \colorbox{green!20}{Scooter Braun's occupation is talent manager and entrepreneur}.}

\textbf{Question 5:} {Who is Pathit Pisitkul?}

\textbf{Response:} {The context is NOT HELPFUL to the question. Based on what I know, \colorbox{green!20}{Pathit Pisitkul is a Thai actor and model}.}

\textbf{Question 6:} {Who is Nicholas Sanduleak?}

\textbf{Response:} {The context is NOT HELPFUL to the question. Based on what I know, \colorbox{green!20}{Nicholas Sanduleak is an astronomer}.}
\end{tcolorbox}}
\caption{Model responses for contradictory and unhelpful contexts, where \colorbox{green!20}{green} indicates model's internal knowledge and \colorbox{blue!20}{blue} indicates context-based information.}
\label{fig:grft_model_responses}
\vspace{-10 pt}
\end{figure*}

\label{sec:appendix}

\end{document}